\pdfoutput=1
\documentclass[letterpaper, 10 pt, conference]{ieeeconf}
\IEEEoverridecommandlockouts
\usepackage{amsmath,amsfonts,amssymb}
\usepackage{algpseudocode}
\usepackage{algorithm}
\usepackage{array}
\usepackage[caption=false,font=small]{subfig}
\usepackage{textcomp}
\usepackage{stfloats}
\usepackage{xurl}
\usepackage{verbatim}
\usepackage{graphicx}
\usepackage{siunitx}
\usepackage{booktabs}
\usepackage{acronym}
\usepackage{multirow}
\usepackage{bm, nicefrac}
\usepackage[colorlinks=true, citecolor=blue, linkcolor=blue, urlcolor=red]{hyperref}
\hypersetup{pdftitle={Precision at Speed: Sample-Efficient Online Model-Based Reinforcement Learning for Hydraulic Excavator Control},pdfauthor={Claudio Canales, Fang Nan, Marco Hutter, Javier Ruiz-del-Solar}}
\usepackage[capitalise]{cleveref}
\usepackage{color, xcolor}

\providecommand{\IEEEpubidadjcol}{}

\acrodef{RL}{Reinforcement Learning}
\acrodef{PPO}{Proximal Policy Optimization}
\acrodef{MBRL}{Model-Based Reinforcement Learning}
\acrodef{MPPI}{Model Predictive Path Integral}
\acrodef{MPC}{Model Predictive Control}
\acrodef{MPCC}{Model Predictive Contouring Control}
\acrodef{SDF}{Signed Distance Field}
\acrodef{MSE}{Mean Squared Error}

\newif\ifanonymous
\anonymousfalse

\DeclareRobustCommand{\anontext}[2]{%
  \ifanonymous
    \textcolor{gray}{#2}
  \else
    #1%
  \fi
}

\begin{document}
\bstctlcite{BSTcontrol}

\title{\LARGE \bf
Precision at Speed: Sample-Efficient Online Model-Based Reinforcement Learning for Hydraulic Excavator Control
}

\author{
\anontext{
Claudio Canales$^{1,2}$, Fang Nan$^{1}$, Marco Hutter$^{1}$, and Javier Ruiz-del-Solar$^{2}$
\thanks{$^{1}$ The authors are with the Robotic Systems Lab, ETH Z\"{u}rich, 8092 Z\"{u}rich, Switzerland.}
\thanks{$^{2}$ The authors are with the Advanced Mining Technology Center (AMTC) and the Department of Electrical Engineering, Universidad de Chile, Santiago, Chile.}
\thanks{Corresponding author: Claudio Canales (e-mail: claudio.canales@amtc.uchile.cl).}
\thanks{This work was supported by  ANID Doctorado Nacional 2023-21232021.}%
\thanks{This work has been submitted to the IEEE for possible publication. Copyright may be transferred without notice, after which this version may no longer be accessible.}%
}
{
Anonymous Authors
}
}

\maketitle
\thispagestyle{empty}
\pagestyle{empty}

\begin{abstract}
Precise, high-speed control remains challenging for robots with complex actuation dynamics. Learning directly on hardware is further constrained by the cost of real-world interaction. We present an online model-based reinforcement learning framework that learns a probabilistic dynamics ensemble model from scratch for sampling-based model predictive control. A precision-gated contouring objective conditions the progress reward on path accuracy, prioritizing precision over speed. In a data-driven excavator simulator, the framework achieves higher sample efficiency than the evaluated model-based reinforcement learning baselines. We validate the framework by learning directly on an 11.5-ton Menzi Muck M445 hydraulic excavator, without demonstrations or simulation pretraining. After 20 minutes of interaction, the controller reaches tracking accuracy comparable to prior learned controllers trained on 100–150 minutes of data. After 40 minutes, it sustains sub-centimeter mean path error at high operating speeds.
\end{abstract}


\section{Introduction}
The control of complex nonlinear dynamical systems is a generally challenging problem.
Traditionally, the control of such systems relies on accurate models of the dynamics, which are often difficult to obtain and require significant manual effort and domain expertise.
During the past decade, data-driven methods have shifted this paradigm, greatly simplifying the modeling process and enabling the control of systems with extremely complex dynamics where analytical modeling is impossible~\cite{Degrave22MagneticControl}.
This trend has been particularly evident in the field of robotics, where learning-based methods have been successfully applied to a wide range of robot control problems.
The field of legged locomotion, for example, has been revolutionized by the use of \ac{RL} methods, which avoid the need for controller design over contact-rich dynamics~\cite{Miki22LearningRobust, Frey26AdvancesChallenges}.

Another field where learning-based methods have shown great promise is the control of robots with complex actuation dynamics, such as hydraulic or pneumatic actuators.
These actuators have advantages in specific applications, but their complex dynamics make them difficult to control with traditional methods.
To address this, learning-based methods have been successfully applied to these systems, including the control of pneumatically actuated robot arms~\cite{Buchler22LearningPlay,Hofer19IterativeLearning}, hydraulic excavators~\cite{Egli22GeneralApproach,Nan24AdaptiveController,leiva2026data}, and cable-driven soft robots~\cite{Nan26EfficientModelbased, Chen24S2C2AFlexible}.
Various learning-based methods have been proposed for these systems, including model-free \ac{RL} methods directly applied to real systems~\cite{Buchler22LearningPlay}, learning a model of the system dynamics and using it as a simulator for \ac{RL} training~\cite{Egli22GeneralApproach}, using a learned model for planning~\cite{Chen24S2C2AFlexible,Zheng25LearningSoft}, and using a learned model for control policy optimization~\cite{Nan26EfficientModelbased}.
\begin{figure}[!t]
\centering
\includegraphics[width=\columnwidth]{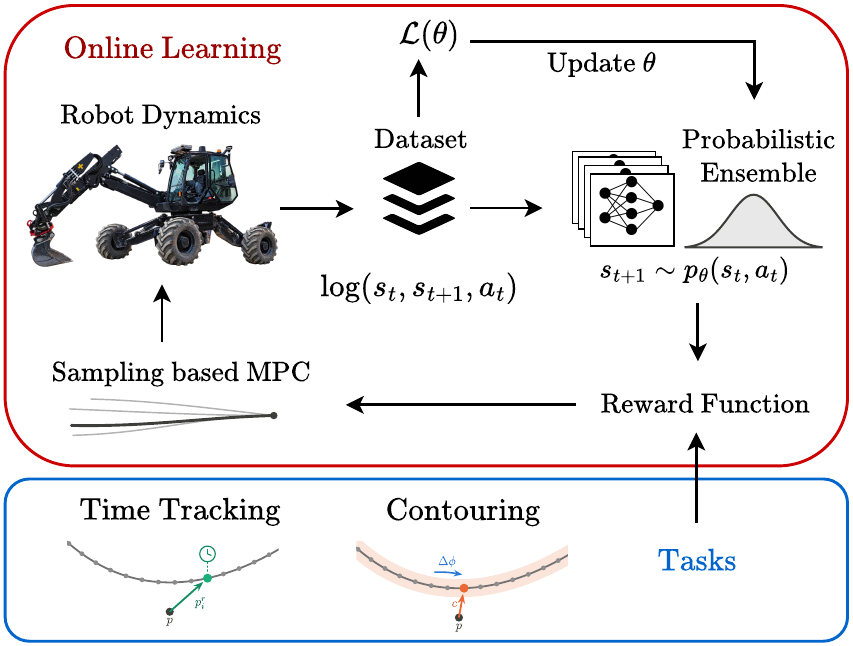}
\caption{Online model-based control framework. Data collected on the machine update a probabilistic dynamics ensemble, through which sampling-based MPC plans under a task-specific reward. The same learned dynamics support time-indexed tracking and contouring.}
\label{fig:method}
\vspace{-4mm}
\end{figure}
\IEEEpubidadjcol

While these methods have shown promising results, sample efficiency, accuracy, and adaptability remain important challenges for real-world deployment, where data collection is expensive, time-consuming, and may expose the hardware to risk.
These challenges are compounded by the heterogeneity of physical systems: even machines of the same class can exhibit different actuation dynamics, demanding sample-efficient methods for learning or fine-tuning controllers on each robot.
In this work, we address these challenges with an online \ac{MBRL} method that learns a probabilistic ensemble of robot dynamics directly from online data collection and uses sampling-based \ac{MPC} for control (Fig.~\ref{fig:method}). The model captures input-dependent variability in the physical dynamics and is learned independently of the task reward. A precision-gated contouring objective prioritizes path accuracy before speed. We demonstrate the complete method on a full-size hydraulic excavator, where it learns precise, high-speed control directly on the machine with limited interaction. The contributions of this work are: 

\begin{itemize}
    \item An online \ac{MBRL} framework that learns a probabilistic dynamics ensemble from scratch through real-world interaction and uses sampling-based \ac{MPC} for control.
    
    \item A precision-gated contouring objective that makes rewarded progress conditional on path accuracy.
    
    \item A real-world validation on an 11.5-ton hydraulic excavator, demonstrating sample-efficient learning and consistent sub-centimeter mean contour error at high operating speed with only 40 minutes of real-world interaction.
\end{itemize}

\section{Related Work}
\label{sec:related}

\subsection{Learning-Based Control of Complex Robotic Systems}

Learning-based control has been applied to systems whose dynamics are dominated by contact, compliant actuation, delays, and coupling, including legged robots~\cite{Miki22LearningRobust,Frey26AdvancesChallenges}, pneumatic and articulated soft robots~\cite{Buchler22LearningPlay,Hofer19IterativeLearning}, and soft robotic arms~\cite{Chen24S2C2AFlexible,Zheng25LearningSoft}. These approaches span model-free control, iterative learning, and planning through learned dynamics, with different requirements for simulation, prior data, and physical interaction.

In hydraulic machinery, an excavator actuator model learned from machine data has been used as a simulator for policy training~\cite{Egli22GeneralApproach}, followed by simulation-trained control with adaptation on the physical machine~\cite{Nan24AdaptiveController}. Sampling-based \ac{MPC} has also been applied to discrete actuation using an offline learned model of a hydraulic impact hammer~\cite{leiva2026data}, while recent work combines online model learning with policy optimization~\cite{Nan26EfficientModelbased}. We instead learn predictive dynamics from autonomously collected machine experience and immediately use them for receding-horizon planning, emphasizing real-machine interaction efficiency and precision at increasing operating speeds.

\subsection{Model-Based Reinforcement Learning}
\ac{MBRL} methods have emerged as a promising approach for sample-efficient learning in robot control tasks.
In contrast to model-free RL, \ac{MBRL} methods learn an explicit model of the task dynamics and use it to generate synthetic data, plan actions, or optimize policies.

Using generated data to improve the model and policy dates back to the Dyna framework~\cite{Sutton90IntegratedModeling} and has been scaled up to high-dimensional tasks with deep neural networks in recent years~\cite{Ha18RecurrentWorld,Hafner25MasteringDiverse}.
Planning through the learned model has also been shown to be effective for robot control tasks, and has been used in various applications~\cite{Nan26EfficientModelbased,Hafner25MasteringDiverse, Hansen22TemporalDifference}.
Recent works have also explored the possibility of using learned models to directly optimize policies, without rolling out the model for imagination data generation~\cite{Nan26EfficientModelbased}.

While a number of works have demonstrated the effectiveness of \ac{MBRL} methods, their application to real-world robot control tasks is still limited due to non-stationary and stochastic dynamics of real-world systems and the time and computational cost for training and planning. We address these issues with a probabilistic dynamics ensemble updated online and a JIT-compiled sampling-based \ac{MPC} pipeline that plans directly through the learned model, and demonstrate real-world feasibility on a full-size hydraulic excavator.

\subsection{\texorpdfstring{\ac{MPCC}}{Model Predictive Contouring Control (MPCC)}}

\ac{MPCC} replaces a time-indexed trajectory with a geometric path and makes traversal pace a decision variable by rewarding progress while penalizing path deviation~\cite{lam2010model,lam2012model}. Conventional formulations introduce a virtual progress state to avoid explicitly projecting each predicted state onto the path; a lag-error penalty couples this state to the system's position along the path. Initially developed for machine-tool control, \ac{MPCC} was later applied to autonomous racing using the track centerline as a progress measure~\cite{liniger2015optimization} and to drone racing using spatially varying contour penalties near gates~\cite{romero2022model}. MPCC++ combines learned aerodynamic corrections with tunnel constraints that tighten near the gates~\cite{krinner2024mpcc++}. In excavation, lifted bucket--soil dynamics enabled convex contouring control in simulation without modeling hydraulic actuator dynamics~\cite{sotiropoulos2021dynamic}. More recently, Zhao et al.~\cite{zhao2025rethinking} showed that sampling-based optimization can accommodate a discontinuous, reference-free gate-progress objective.

We instead use sampling-based \ac{MPC} for precision-prioritized path following. Each rollout state is projected onto a discrete forward reference window, yielding the exact discrete projection and its associated progress without an optimized virtual progress state or lag-error penalty. A multiplicative error gate attenuates rewarded progress as contour error increases, encoding path accuracy as the priority over traversal speed.

\section{Method}
\label{sec:method}
The proposed method couples online dynamics learning with receding-horizon control directly on the physical system (Fig.~\ref{fig:method}). Each applied action generates experience; after each episode, the accumulated data update a probabilistic dynamics ensemble, which \ac{MPPI} uses to plan actions in every control cycle. Because the learned dynamics are independent of the reward, the same model supports time-indexed tracking, pace-capped contouring, and precision-gated free contouring. The following sections formalize the control problem, introduce the model and planner, define the control objectives, and explain how these components are integrated into a real-time online learning loop deployed on the physical excavator.

\subsection{Problem Setting}
\label{sec:method:setting}

We consider a nonlinear plant whose state \(s_t\) evolves under action \(a_t\) as
\begin{equation}
\label{eq:plant}
s_{t+1}=f(s_t,a_t)+w_t,
\end{equation}
where the transition map \(f\) and process noise \(w_t\) are unknown. From the transitions collected online, we learn \(p_\theta(s_{t+1}\mid s_t,a_t)\) and use it for receding-horizon control.

We instantiate this problem on a hydraulic excavator, controlling the boom, stick, telescopic extension, and shovel pitch (Fig.~\ref{fig:m445}). Actions are current setpoints to the electric pilot-stage valves, observations are joint positions and velocities, and the task is to follow end-effector paths in the cabin frame. The pilot and main hydraulic stages introduce input delays, dead zones, load-dependent nonlinearities, and coupling between actuators. We therefore learn the dynamics entirely through online interaction with the machine, without hydraulic-specific knowledge, offline data, simulation pretraining, or demonstrations.
\subsection{Probabilistic Model}
\label{sec:method:model}

Within the general formulation of Eq.~\eqref{eq:plant}, the state is constructed from the signals available on the controlled system. On the excavator, joint positions $q_t$ and velocities $\dot q_t$ are measured, and applied valve commands $a_t$ are recorded, whereas hydraulic pressures and internal valve states remain unobserved. To represent their delayed influence on the joint response, we define
\begin{equation}
\label{eq:state}
s_t =
\left(
q_{t-k:t},\;
\dot q_{t-k:t},\;
a_{t-k:t-1}
\right),
\end{equation}
where \(k\) denotes the history horizon.
This finite history provides the temporal context needed to represent the delayed hydraulic response using measured signals alone, without direct pressure sensing or an explicitly identified delay model.

We represent the transition with a probabilistic ensemble of feed-forward networks, following the model class used in PETS~\cite{chua2018deep}. Suppressing the member index, each network predicts a diagonal Gaussian distribution over the next joint velocity, from which the position follows by integration:
\begin{equation}
\label{eq:model}
\begin{aligned}
\dot q_{t+1}
&\sim
\mathcal{N}\!\left(
\mu_{\theta}(s_t,a_t),
\Sigma_{\theta}(s_t,a_t)
\right), \\
q_{t+1}
&=q_t+\dot q_{t+1}\Delta t,
\end{aligned}
\end{equation}
where
\(\Sigma_{\theta}=\operatorname{diag}(\sigma_{\theta}^{2})\).
Predicting velocity avoids learning absolute joint offsets, while integration enforces consistency between predicted positions and velocities. The predicted variance models input-dependent noise, whereas disagreement among member means provides a proxy for model uncertainty.

After each episode, all accumulated data are used to update the model through open-loop rollouts of length \(H\). Starting from a logged state \(\hat{s}_0\), predicted states are recursively fed back under the recorded actions, and the measured velocities are evaluated under the predicted distributions:
\begin{equation}
\label{eq:loss}
\mathcal{L}(\theta)
=
-\frac{1}{H}
\sum_{i=0}^{H-1}
\log
\mathcal{N}\!\left(
\dot q^{*}_{i+1};
\mu_{\theta}(\hat{s}_i,a_i),
\Sigma_{\theta}(\hat{s}_i,a_i)
\right),
\end{equation}
where \(\hat{s}_i\) is generated recursively by the model. The loss is evaluated for each ensemble member and averaged during training. It balances prediction error against predicted variance, while backpropagation through the rollout penalizes compounding error over the training rollout horizon.

We retain the probabilistic ensemble model of PETS but adapt its training and propagation to onboard control: the model is trained through multistep rollouts rather than one-step transitions, and the planner propagates the ensemble-mean prediction instead of sampled particles. Multistep training addresses the recursive prediction error, while mean propagation reduces the computational cost in real-time control.
\subsection{Planning with the Learned Model}
\label{sec:method:planner}

In each control cycle, the planner maximizes the task reward \(r\) of Sec.~\ref{sec:method:objectives} over a horizon of \(T\) steps by propagating the mean learned transition:
\begin{equation}
\label{eq:planning}
\begin{aligned}
a^{*}_{0:T-1}
&=
\arg\max_{a_{0:T-1}}
\sum_{i=0}^{T-1} r(\bar{s}_i,a_i), \\
\mathrm{s.t.}\qquad
\bar{s}_0 &= s_t, \qquad
\bar{s}_{i+1}
=
\mathbb{E}_{p_\theta}
\!\left[s_{i+1}\mid\bar{s}_i,a_i\right].
\end{aligned}
\end{equation}
Only the first optimized action is applied before replanning from the next measurement. We solve Eq.~\eqref{eq:planning} using \ac{MPPI}~\cite{Williams17InformationTheoretic}, which evaluates sampled action sequences without requiring reward gradients. The sampled sequences and executed commands use the same per-joint exponential moving-average filter, ensuring consistency between planned and applied actions.

\subsection{Reward Functions}
\label{sec:method:objectives}
The reward function in Eq.~\eqref{eq:planning} specifies the control task independently of the learned dynamics. A task is defined by a reference joint path \(q^r_{0:M}\) and its corresponding end-effector path \(p_m^r=\mathrm{FK}(q_m^r)\), where \(M\) is the number of reference points. We consider time-indexed tracking and contouring.

\subsubsection{Time-Indexed Tracking}
Time-indexed tracking couples the path to its prescribed schedule. At rollout step \(i\), the state is compared with the \(i\)-th reference point:
\begin{equation}
\label{eq:reward:track}
r^{\mathrm{tr}}_i
=
-w_q\left\lVert q_i-q_i^r\right\rVert^2
-w_p\left\lVert p(q_i)-p_i^r\right\rVert^2
-w_a\left\lVert\Delta a_i\right\rVert^2,
\end{equation}
where \(w_q\) and \(w_p\) weight the joint and end-effector tracking errors, and \(\Delta a_i=a_i-a_{i-1}\) is the action increment. The action-rate penalty \(w_a\lVert\Delta a_i\rVert^2\) discourages abrupt changes in the valve commands. The controller must therefore follow both the path geometry and its timing.

\begin{figure}[t]
\centering
\includegraphics[width=\columnwidth]{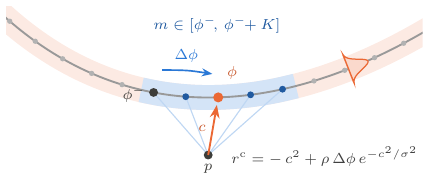}
\caption{Contouring. The current state is projected onto the forward reference window \(m\in[\phi^{-},\phi^{-}+K]\) (blue). The minimizer \(\phi\) defines the weighted contour cost \(c\) and normalized progress \(\Delta\phi\); the precision gate attenuates rewarded progress as \(c\) increases.}
\label{fig:contouring}
\end{figure}

\subsubsection{Contouring}
Contouring removes the prescribed timing by comparing the state with its projection onto the reference (Fig.~\ref{fig:contouring}). Given the previous path index \(\phi^{-}\), the weighted squared discrepancy to each point in the forward window is
\begin{equation}
\label{eq:reward:proj}
\begin{aligned}
d_m &= w_q\left\lVert q_i-q_m^r\right\rVert^2+
w_p\left\lVert p(q_i)-p_m^r\right\rVert^2, \\
&\qquad m\in[ \phi^{-},\phi^{-}+K].
\end{aligned}
\end{equation}
The projection \(\phi=\arg\min_m d_m\) defines the weighted contour measure \(c^2=d_\phi\) and normalized progress \(\Delta\phi=(\phi-\phi^{-})/M\). Because the search begins at \(\phi^{-}\), the projection is monotonic and can advance by at most \(K\) reference points per control cycle. The reward is
\begin{equation}
\label{eq:reward:cont}
\begin{split}
r^{\mathrm{c}}_i
={}&-c^2+\rho\,\Delta\phi\,e^{-c^2/\sigma^2} \\
&-w_v\sum_j\max\!\left(0,|\dot q_j|-\bar v_j\right)^2
-w_a\left\lVert\Delta a_i\right\rVert^2,
\end{split}
\end{equation}
where \(\rho\) weights progress, \(\sigma\) sets the decay scale of the precision gate, and \(w_v\) imposes soft joint-speed limits at \(\bar v_j\). Time-indexed tracking requires no corresponding speed penalty because its schedule already regulates the operating pace. The gate makes progress conditional on accuracy: precision is prioritized, and speed is rewarded only where that precision is maintained.

The window size \(K\) determines the admissible pace. Because the reference is sampled at the control rate, \(K=1\) limits progress to the reference pace while removing explicit time indexing, whereas larger \(K\) permits faster progress. The experiments compare time-indexed tracking, pace-capped contouring (\(K=1\)), and free contouring (\(K=7\)).
\subsection{Online Learning Loop}
\label{sec:method:loop}

\begin{algorithm}[b]
\caption{Online model-based learning on the machine}
\label{alg:loop}
\footnotesize
\begin{algorithmic}[1]
\Require reward \(r\), episodes \(E\), trajectories per episode \(n\)
\State \(\mathcal{D}\gets\textsc{WarmStart}()\)
       \Comment{fixed-amplitude random sinusoids}
\State \(\theta\gets\textsc{Train}(\mathcal{D})\)
       \Comment{Eq.~\eqref{eq:loss}}
\For{episode \(e=1,\ldots,E\)}
    \State \(\beta_e\gets\) action limit;
           \(\mathcal{D}_e\gets\varnothing\)
    \For{trajectory \(j=1,\ldots,n\)}
        \State \(q^\star\gets\) sample target uniformly in joint box
        \State \(q^r_{0:M}\gets\) quintic reference from \(q\) to \(q^\star\)
        \Repeat
            \State \(s_t\gets\) current measurement history
            \State \(a^*_{0:T-1}\gets
                   \textsc{MPPI}(s_t,\theta,r)\)
            \State \(\tilde a_t\gets
                   \operatorname{EMA}(a^*_0,a_{t-1})\)
            \State \(a_t\gets\tilde a_t\);
                   clip to \([-\beta_e,\beta_e]\);
                   apply joint-limit barrier
            \State apply \(a_t\) and observe \(s_{t+1}\)
            \State \(\mathcal{D}_e\gets
                   \mathcal{D}_e\cup\{(s_t,a_t,s_{t+1})\}\)
        \Until{trajectory terminates}
    \EndFor
    \State \(\mathcal{D}\gets\mathcal{D}\cup\mathcal{D}_e\)
    \State \(\theta\gets\textsc{Train}(\mathcal{D})\);
           checkpoint \((\theta,\mathcal{D})\)
\EndFor
\end{algorithmic}
\end{algorithm}

Online model-based control presents a bootstrap problem: planning requires a dynamics model, but the model can only be learned from actions applied to the robot. 
The loop in Fig.~\ref{fig:method}, summarized in Algorithm~\ref{alg:loop}, resolves this dependency directly on the excavator.
Before enabling the planner, bounded sinusoidal valve-current commands with randomized periods and phases collect the initial dataset.
A directional joint-limit barrier suppresses command components pointing outside the admissible joint region, and the resulting data are used to train the first ensemble. Once model-based control begins, we bound each joint's normalized valve-current command to $[-\beta_e,\beta_e]$ during episode $e$. To limit aggressive commands during early learning, the bound starts at $\beta_1=0.5$ and increases by $0.1$ per episode to a maximum of $1.0$ (Table~\ref{tab:params}).

Learning then proceeds in episodes of consecutive trajectories. For each trajectory, a target is sampled uniformly from a box in joint space, and a minimum-jerk quintic reference is generated from the current configuration. In every control cycle, the measurement history forms $s_t$, and \ac{MPPI} plans through the current model with the selected reward. The first planned action $a_0^*$ is smoothed using the per-joint exponential moving average (EMA), $ \tilde{a}_t = \alpha a_0^* + (1-\alpha)a_{t-1},$
where \(a_{t-1}\) is the command previously applied. The same filter is included in the sampled rollouts so that the planner evaluates the commands received by the machine. The filtered command \(\tilde a_t\) is clipped component-wise to \([-\beta_e,\beta_e]\) and passes through the joint-limit barrier, yielding the applied action \(a_t\). After applying \(a_t\) to the pilot-stage valves, the transition \((s_t,a_t,s_{t+1})\) is recorded. Time-indexed trajectories follow the complete reference schedule, whereas contouring trajectories terminate upon path completion or timeout.
The model remains fixed while an episode collects data. After the episode, the new transitions are appended to the accumulated dataset, and the ensemble is updated using Eq.~\eqref{eq:loss}. The updated model is deployed in the next episode. Thus, planning and data collection are run at the control rate, whereas model updates occur between episodes.

\section{Simulation Results}
\label{sec:sim}

To evaluate the proposed method, we conduct simulation experiments on a data-driven simulator of a hydraulic excavator~\cite{Egli22GeneralApproach}, which has been used as a benchmark for model-based reinforcement learning methods~\cite{Nan26EfficientModelbased}.
We compare the sample efficiency of our method against various \ac{MBRL} baselines.

\subsection{Simulation Setup}
\label{sec:sim:setup}
The simulation environment uses a learned model of a hydraulic excavator~\cite{Egli22GeneralApproach}, represented as a neural network that predicts the joint velocities after one control step given the current state and the control input.
Previous research has shown that this model resembles the real system well enough that the methods validated with this model transfer to the real system~\cite{ Egli22GeneralApproach, Nan24AdaptiveController, Nan26EfficientModelbased, Egli20RLBasedHydraulic}.
We apply the proposed method to the same task of tracking arbitrary polynomial trajectories used in~\cite{Nan26EfficientModelbased}.
All methods follow the benchmark protocol of~\cite{Nan26EfficientModelbased} with identical environment-interaction budgets; the baselines are rerun using their official implementations, and the curves in Fig.~\ref{fig:sim_baseline_training} report the mean and standard deviation over three seeds per method.

\subsection{Sample Efficiency Against Model-Based Baselines}
\label{sec:sim:baselines}
\begin{figure}
  \centering
  \includegraphics[width=0.9\columnwidth, trim={6pt 6pt 0pt 6pt},clip]{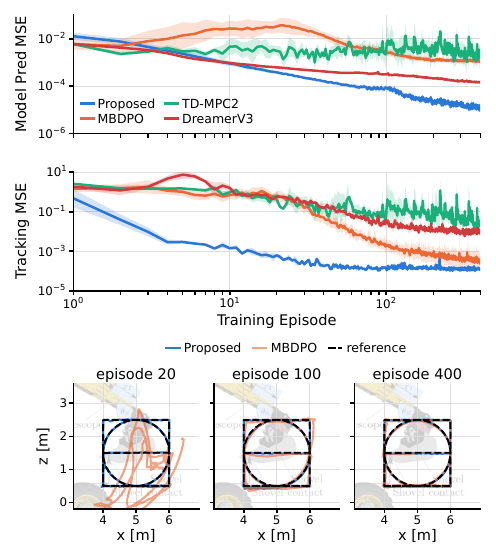}
  \caption{Top: one-step model MSE. Middle: tracking MSE. Bottom: end-effector trajectories after 20, 100, and 400 training episodes. For the latent-space models, we plot the latent-space prediction MSE loss of TD-MPC2 and the decoded state-space prediction MSE of Dreamer, as the latter method uses categorical latent variables in its open-source implementation. The model errors are scaled based on their value at iteration 0 to normalize the scale of different prediction spaces.}
  \label{fig:sim_baseline_training}
\end{figure}

Figure~\ref{fig:sim_baseline_training} shows the loss of model prediction and control performance of the proposed method compared to various model-based baselines, including TD-MPC2~\cite{Hansen24TDMPC2Scalable}, DreamerV3~\cite{Hafner25MasteringDiverse}, and model-based differentiable policy optimization (MBDPO)~\cite{Nan26EfficientModelbased}.
All baselines learn dynamics models of the system, and use the learned models for planning, generating data for policy optimization, and calculating approximate policy gradients, respectively.

The proposed method achieves significantly better sample efficiency than all baselines.
The model performance shown in~\cref{fig:sim_baseline_training} is the \ac{MSE} in single-step prediction of the learned model.
While the proposed method optimizes the model with the negative log-likelihood loss, it still outperforms all baselines that are directly trained to minimize prediction \ac{MSE}.
Additionally, the proposed method achieves the control performance reported in~\cite{Nan26EfficientModelbased} with only 20 episodes, while the baseline requires more than 100.

We hypothesize that the advantage of the proposed method comes from the use of a probabilistic ensemble of dynamics models, which explicitly models the aleatoric uncertainty of the dynamics.
On the contrary, the baselines, using model architectures lacking such expressiveness, struggle to learn stochastic dynamics from limited data, leading to poor generalization and limited control performance.
Unlike MBDPO's deterministic ensemble, our probabilistic ensemble explicitly models aleatoric noise and uses the predicted variance to down-weight noisy transitions during training.
We also note that the EMA filter on the control actions forces the controller to act smoothly and stay consistent between replans.
This avoids high-frequency actions being executed on the system, improving controller performance as well as avoiding model overfitting on oscillatory data, which is a severe problem for the baseline methods.

The trajectory tracking performance of the proposed method and the MBDPO baseline is shown in~\cref{fig:sim_baseline_training}.
We plot the end-effector trajectories of the excavator at different training episodes of the proposed method and MBDPO, the strongest baseline among the three.
It is observed that the proposed method achieves better trajectory tracking performance at episode 20, while the MBDPO baseline still behaves randomly.
At episode 100, the proposed method achieves near-perfect tracking performance, while the MBDPO baseline still has noticeable tracking error.
This comparison highlights the superior sample efficiency of the proposed method, suggesting that it may be more suitable for real-world applications where data collection is expensive and time-consuming.

\section{Real-World Experiments}
\begin{figure}[!t]
\centering
\includegraphics[width=0.8\columnwidth]{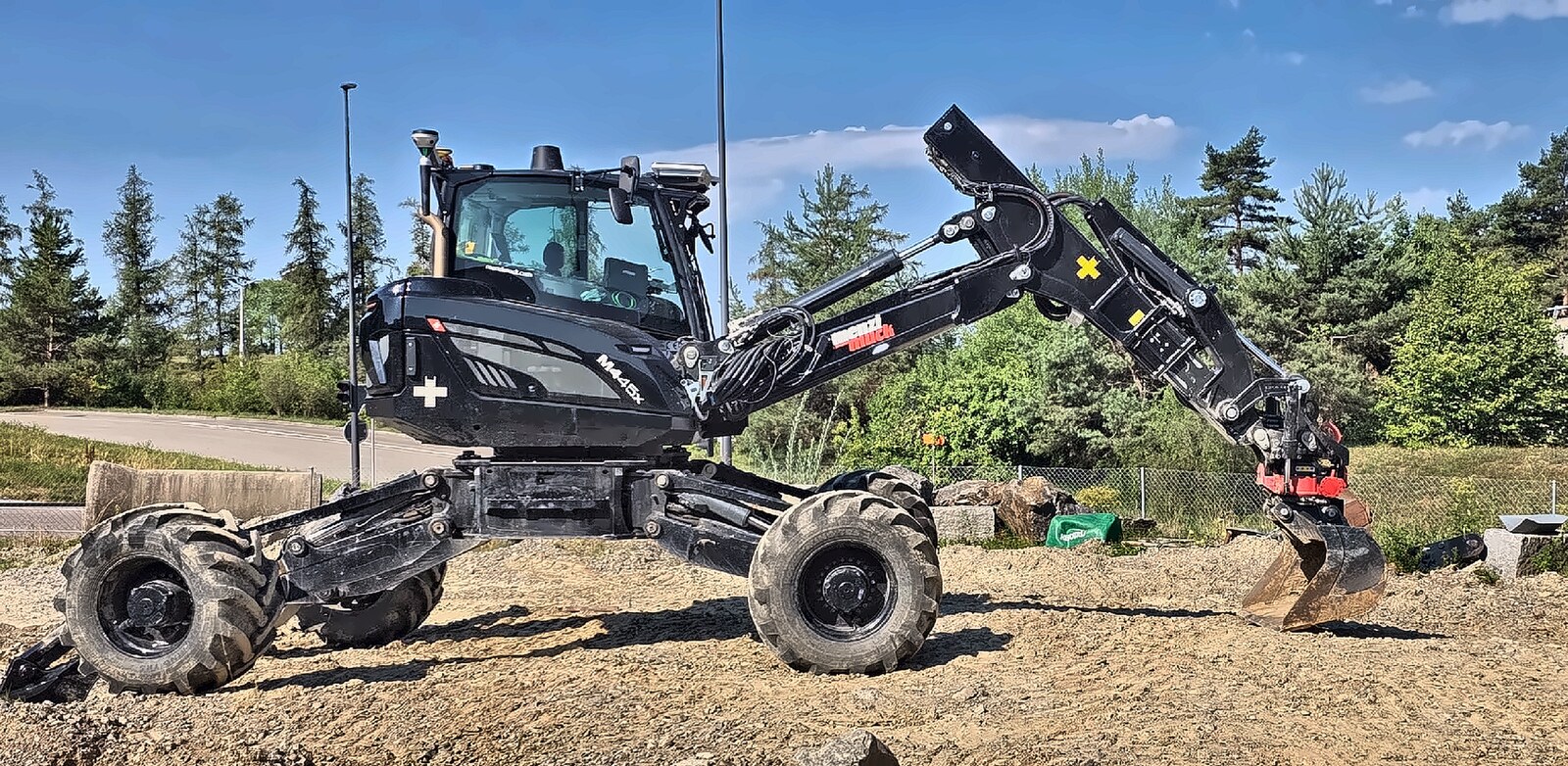}
\caption{Experimental platform: the 11.5-ton Menzi Muck M445 hydraulic excavator, whose four arm joints are controlled through the learned dynamics model.}
\label{fig:m445}
\vspace{-4mm}
\end{figure}

\begin{figure*}[!t]
\centering
\includegraphics[width=0.88\textwidth, trim={6pt 4pt 6pt 8pt},
  clip]{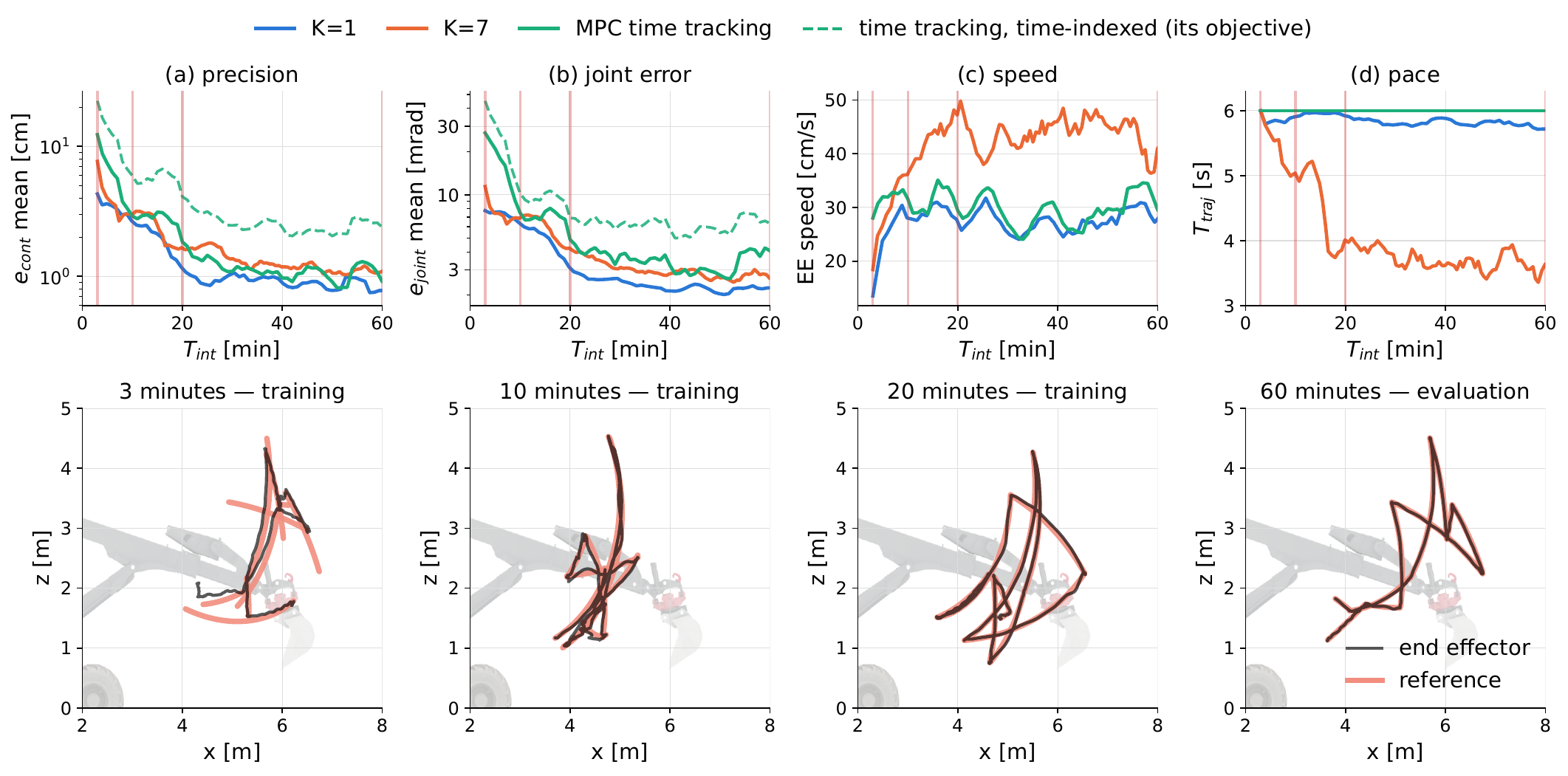}

\caption{Training from scratch on the machine, one run per reward mode.
Top: (a) contour error $e_{cont}$, (b) joint contour error $e_{joint}$,
(c) end-effector speed, (d) time per trajectory $T_{traj}$, over
interaction time $T_{int}$; 5-episode rolling means; dashed: time
tracking under its own time-indexed objective; red marks: the snapshots
below. Bottom: $K{=}7$ end-effector paths at 3, 10, and 20 minutes of
training and at the frozen 60-minute evaluation.}
\label{fig:learning}
\vspace{-4mm}  
\end{figure*}
\label{sec:experiments}
We evaluate the methodology through three one-hour online learning experiments conducted from scratch on an 11.5-ton Menzi Muck M445, without a prior dynamics model. One run is performed under each objective: time-indexed tracking, pace-capped contouring (\(K=1\)), and free contouring (\(K=7\)). Experience is collected by executing minimum-jerk quintic trajectories between targets sampled across the joint workspace. The experiments evaluate: (i) how efficiently and precisely the method learns on a full-scale excavator relative to prior learned controllers; (ii) whether error-gated progress enables higher speeds without sacrificing path accuracy; and (iii) whether learned dynamics transfer across deployment objectives and how performance depends on the training objective. We examine these questions through the online learning results, a controlled gate ablation, and a \(3\times3\) cross-evaluation of the trained models and objectives.

\subsection{Setup and Metrics}
\label{sec:exp:setup}

\begin{table}[!t]
\caption{System, model, planner, and reward parameters.}
\label{tab:params}
\centering
\scriptsize
\setlength{\tabcolsep}{2pt}
\renewcommand{\arraystretch}{0.95}
\begin{tabular}{@{}cll@{\hspace{6pt}}cll@{}}
\toprule
\multirow{7}{*}{\rotatebox[origin=c]{90}{\textit{System}}}
 & Control rate & \SI{25}{Hz} &
\multirow{7}{*}{\rotatebox[origin=c]{90}{\textit{Planner (\acs{MPPI})}}}
 & Samples $N$ & 3000 \\
 & Controlled joints & 4 & & Horizon $T$ & 30 (\SI{1.2}{s}) \\
 & Command & valve current & & Iterations & 3 \\
 & Reference & \SI{6}{s} quintic & & Temperature $\lambda$ & 0.05 \\
 & Trajectories/episode & 10 & & Sampling std & 0.5 \\
 & Warm start & $2 \times \SI{60}{s}$ & & EMA $\alpha$ & 0.18 \\
 & Authority $\beta_e$ & $0.5 \rightarrow 1.0$, $+0.1$/ep. & & & \\
\midrule
\multirow{6}{*}{\rotatebox[origin=c]{90}{\textit{Model}}}
 & Ensemble members & 5 &
\multirow{6}{*}{\rotatebox[origin=c]{90}{\textit{Rewards}}}
 & $w_q$, $w_p$ & 8, 2 \\
 & Hidden layers & $2 \times 256$ & & $\rho$, $\sigma$ & 20, 0.05 \\
 & History $k$ & 15 (\SI{0.6}{s}) & & $K$ & 1 or 7 \\
 & Rollout loss $H$ & 10 & & $\bar{v}$ & \SI{0.6}{rad/s} \\
 & Optimizer, lr & Adam, $10^{-4}$ & & $w_v$, $w_a$ & 50, 0.05 \\
 & Batch, epochs/ep. & 128, 3 & & Eval timeout & \SI{8}{s} \\
\bottomrule
\end{tabular}
\vspace{-4mm}
\end{table}

All real-world experiments use a Menzi Muck M445 (Fig.~\ref{fig:m445}), controlling the boom, stick, telescopic extension, and shovel pitch through current commands to the electric pilot-stage valves using measured joint positions and velocities. The JIT-compiled JAX implementation runs onboard on an NVIDIA RTX 5080 GPU and AMD Ryzen 9 7900X CPU. Actions are updated every \SI{40}{ms}, with each \ac{MPPI} update taking \SI{7.6}{ms} on average and \SI{8.1}{ms} at most. All experiments share the model architecture, planner configuration, and action filtering given in Table~\ref{tab:params}; only the deployed reward changes.

Geometric accuracy is measured at each control cycle by the end-effector contour error \(e_{\mathrm{cont}}\), the Euclidean distance to the closest point on the reference path. For time-indexed tracking, \(e_{\mathrm{time}}\) additionally measures the distance to the scheduled reference point, separating path geometry from timing. We report the mean, 95th percentile, and maximum errors, together with mean and maximum end-effector speed. Cross-study comparisons use the speed-normalized maximum error \(\rho_M=|e|_{\max}/v_{\max}\)~\cite{Mattila17SurveyControl}, where lower values indicate greater precision relative to speed. Sample efficiency is measured separately by real-machine interaction time \(T_{\mathrm{int}}\), with simulation pretraining reported where applicable.

\subsection{Online Learning Performance}
\label{sec:exp:learning}
We assess online learning performance using both the behavior observed during data collection and controlled evaluations of the learned models. Figure~\ref{fig:learning} reports episode-level contour error, joint error, end-effector speed, and trajectory duration for one training run under each objective, together with representative executed paths at selected interaction times. From each run, we freeze the models obtained after 20, 40, and 60 minutes and evaluate them without further updates on matched seeded sequences of target trajectories. Table~\ref{tab:sota} reports these checkpoint evaluations. The training results show how control evolves as the model collects experience, while the frozen evaluations provide a consistent basis for comparing performance across interaction times and with prior work. 

The training curves reveal a common pattern across the experiments: the largest improvement in control accuracy occurs during the first 20 minutes of interaction. Both the end-effector and joint contour errors decrease substantially as the controller collects data and updates its dynamics model (Fig.~\ref{fig:learning}a,b). The \(K=7\) trajectory snapshots illustrate the same progression: large deviations from the reference are visible after three minutes, the path geometry is followed more closely after ten minutes, and consistent path execution is observed after twenty minutes. The frozen 20-minute evaluations confirm this improvement, with mean contour errors between \SI{0.87}{cm} and \SI{1.4}{cm} across the three objectives (Table~\ref{tab:sota}).

This common improvement in accuracy is accompanied by a clear separation in operating pace. Time-indexed tracking and pace-capped contouring (\(K=1\)) remain close to the pace prescribed by the reference, whereas free contouring (\(K=7\)) settles at approximately one and a half times their mean end-effector speed and reduces the mean trajectory duration by about one third (Fig.~\ref{fig:learning}c,d). This separation develops within the first 20 minutes and persists throughout the remainder of training, while the mean contour errors remain near the centimeter level.

Table~\ref{tab:sota} compares our results with the closest learned controllers, evaluated on the slightly larger 12.5-ton M545~\cite{Egli22GeneralApproach,Nan24AdaptiveController,Nan26EfficientModelbased}. Because prior studies report time-indexed error, whereas contour error ignores timing and is roughly half as large in our tracking runs, we compare \(\rho_M\) only within the same convention; contour-based values are marked \(^\dagger\). Using time-indexed error, our controller reaches the range of prior results after 20 minutes, compared with 100--150 minutes of machine data and, in two cases, simulation pretraining. After 40 minutes, all objectives achieve sub-centimeter mean contour error, while both contouring modes reach and maintain \(\rho_M=\SI{0.04}{s}\), equivalent to the distance traveled during one control update at peak speed.

To evaluate sustained high-speed operation, we test the 60-minute \(K{=}7\) controller on paths \(2.7\) times longer than in the standard evaluation. It maintains \SI{0.69}{cm} mean contour error while averaging \SI{75}{cm/s} and reaching \SI{180}{cm/s}, above the \SI{129}{cm/s} previously reported on the M545~\cite{Nan26EfficientModelbased}, with a speed-normalized error of \SI{0.02}{s}.

\begin{table}[t]
\caption{Comparison with learned excavator controllers; $\rho_M = |e|_{max} / v_{max}$ is the speed-normalized error metric.}
\label{tab:sota}
\centering
\footnotesize
\setlength{\tabcolsep}{1.35pt}
\renewcommand{\arraystretch}{1.12}
\begin{tabular}{@{}c l c cc ccc ccc c@{}}
\toprule
 & & $\rho_M$ & \multicolumn{2}{c}{$v$ [cm/s]} & \multicolumn{3}{c}{$e_{time}$ [cm]} & \multicolumn{3}{c}{$e_{cont}$ [cm]} & $T_{int}$ \\
\cmidrule(lr){4-5} \cmidrule(lr){6-8} \cmidrule(lr){9-11}
 & & [s] & avg & max & avg & p95 & max & avg & p95 & max & [min] \\
\midrule
\multirow{3}{*}{\rotatebox[origin=c]{90}{\textbf{M545}}}
 & Egli \& Hutter~\cite{Egli22GeneralApproach} & 0.13 & 20 & 53 & 2.8 & -- & 6.7 & -- & -- & -- & 100 \\
 & Nan \& Hutter~\cite{Nan24AdaptiveController} & 0.27 & 37.5 & 60 & 8.56 & -- & 16.2 & -- & -- & -- & sim+10 \\
 & Nan et al.~\cite{Nan26EfficientModelbased} & 0.10 & 52 & 129 & 4.5 & -- & 12.4 & -- & -- & -- & 150 \\
\midrule
\multirow{10}{*}{\rotatebox[origin=c]{90}{\textbf{Ours, M445}}}
 & Time tracking, 20 & 0.12 & 28 & 137 & 2.7 & 7.9 & 16.3 & 1.1 & 3.1 & 7.5 & 20 \\
 & $K{=}1$, 20 & 0.05$^\dagger$ & 26 & 94 & -- & -- & -- & 0.87 & 2.5 & 5.1 & 20 \\
 & $K{=}7$, 20 & 0.11$^\dagger$ & 43 & 147 & -- & -- & -- & 1.4 & 3.9 & 15.5 & 20 \\
\cmidrule(lr){2-12}
 & Time tracking, 40 & 0.13 & 26 & 94 & 2.2 & 5.3 & 11.9 & 0.98 & 2.8 & 5.0 & 40 \\
 & $K{=}1$, 40 & 0.04$^\dagger$ & 25 & 99 & -- & -- & -- & 0.72 & 1.9 & 4.4 & 40 \\
 & $K{=}7$, 40 & 0.04$^\dagger$ & 42 & 152 & -- & -- & -- & 0.93 & 2.7 & 5.5 & 40 \\
\cmidrule(lr){2-12}
 & Time tracking, 60 & 0.12 & 27 & 82 & 2.0 & 5.1 & 9.9 & 0.72 & 2.2 & 5.3 & 60 \\
 & $K{=}1$, 60 & 0.04$^\dagger$ & 25 & 99 & -- & -- & -- & 0.75 & 2.1 & 3.9 & 60 \\
 & $K{=}7$, 60 & 0.04$^\dagger$ & 44 & 139 & -- & -- & -- & 0.82 & 2.3 & 5.0 & 60 \\
 & Long paths, 60 & 0.02$^\dagger$ & 75 & 180 & -- & -- & -- & 0.69 & 2.0 & 4.0 & 60 \\
\bottomrule
\multicolumn{12}{@{}l@{}}{\scriptsize $^\dagger$\,$\rho_M$ computed from $e_{cont}$; no schedule exists for contouring.}
\end{tabular}
\end{table}

\subsection{Precision at Speed}
\label{sec:exp:pareto}

\begin{figure}[!t]
\centering
\includegraphics[width=0.85\columnwidth, trim={7pt 6pt 0pt 7pt}, clip]{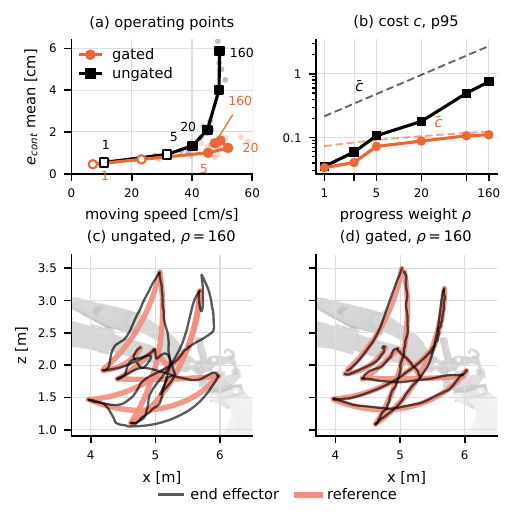}
\caption{Precision-gate ablation with the frozen 60-minute time-tracking model (\(\sigma=0.05\); auxiliary penalties disabled). (a) Mean contour error versus end-effector speed across \(\rho\); open markers indicate below 90\% completion. (b) Measured p95 contour cost and theoretical break-even thresholds without fitting. (c),(d) Ungated and gated paths at \(\rho=160\).}
\label{fig:gate}
\end{figure}

Conventional MPCC uses an ungated progress reward, whereas our formulation makes progress conditional on contour accuracy. With all auxiliary reward penalties removed, the two objectives are
\begin{equation}
\label{eq:gate:objectives}
\begin{aligned}
r_{\mathrm{ungated}}
    &= -c^2 + \rho\Delta\phi, \\
r_{\mathrm{gated}}
    &= -c^2 + \rho\Delta\phi
       \exp\!\left(-\frac{c^2}{\sigma^2}\right),
\end{aligned}
\end{equation}
where \(c\) is the contour cost combining end-effector and joint-position errors. Accordingly, \(\sigma=0.05\) defines a scale in this mixed cost rather than a distance in meters. We compare both objectives using the same frozen 60-minute time-tracking model, \(K=7\), and a sweep of \(\rho\) spanning two orders of magnitude (Fig.~\ref{fig:gate}). Each configuration is evaluated using three matched seeds with ten consecutive trajectories per seed. The learned model, planner, target-generation protocol, and contour penalty are therefore fixed, isolating the effect of the gate.

Both objectives increase the operating speed as \(\rho\) grows and eventually approach a similar speed range (Fig.~\ref{fig:gate}a). At low \(\rho\), both controllers remain precise at low speeds. Their behavior diverges as the progress incentive increases: the ungated controller purchases progress through increasing contour error, whereas the gated controller increases speed while keeping the error within a narrow region. The trajectories at \(\rho=160\) show the resulting difference (Fig.~\ref{fig:gate}c,d): the ungated controller cuts across the reference paths, while the gated controller follows their geometry at a comparable operating speed.

The error at which progress ceases to pay off is obtained by setting each one-step reward in~\eqref{eq:gate:objectives} to zero at the maximum progress, \(\Delta\phi=K/M\), giving
\begin{equation}
\label{eq:gate:bounds}
\begin{aligned}
\bar c_{\mathrm{ungated}}
    &= \sqrt{\frac{\rho K}{M}}, \quad \bar c_{\mathrm{gated}}
    &= \sigma\sqrt{
       W\!\left(\frac{\rho K}{M\sigma^2}\right)}
\end{aligned}
\end{equation}
where \(W\) is the Lambert \(W\) function. Above \(\bar c\), even the maximum possible progress cannot compensate for the contour penalty. The gate therefore replaces the classical \(\sqrt{\rho}\) accuracy--progress trade-off with a precision-prioritized relation whose break-even error grows only as \(\sqrt{W(\rho)}\), approximately \(\sqrt{\log\rho}\). Figure~\ref{fig:gate}b shows that the measured 95th-percentile weighted contour costs follow the respective bounds in~\eqref{eq:gate:bounds} without fitting them to the data. Over the tested range, this keeps the contour cost within a narrow region while allowing speed to increase and saturate: precision is established first, and progress is rewarded only where that precision can be maintained. The real-machine results follow this predicted behavior: the ungated controller exchanges accuracy for progress as \(\rho\) increases, whereas the gated controller reaches a comparable speed while maintaining a bounded contour cost.
\subsection{Task Dependence}
\label{sec:exp:matrix}

\begin{figure}[t]
\centering
\includegraphics[width=0.88\columnwidth, trim={6pt 6pt 0pt 6pt}, clip]{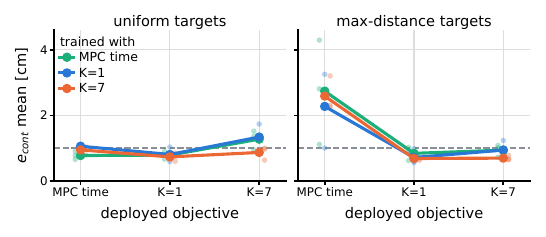}
\caption{$3{\times}3$ cross-evaluation at 60 minutes (lines: trained model;
dots: seed means). The deployed objective sets performance; the models
separate only under the $K{=}7$ evaluation.}
\label{fig:crosseval}
\end{figure}

The preceding experiments pair each model with its training objective, confounding the effects of the learned dynamics and the deployed objective. We separate them through a \(3\times3\) cross-evaluation in which each frozen 60-minute model is deployed under all three objectives. Every pairing is evaluated on uniform and maximum-distance targets using matched seeds and an unchanged planner, testing whether the learned dynamics support objectives not used during data collection.
Figure~\ref{fig:crosseval} shows that accuracy depends primarily on the deployed objective. All nine pairings remain near one-centimeter mean contour error on uniform targets. On maximum-distance targets, time-indexed tracking incurs greater error under the faster schedule, whereas both contouring objectives retain sub-centimeter means. Model differences appear only under \(K=7\), where the model trained at that pace is slightly more accurate. Thus, the learned dynamics transfer across the objectives considered, with modest model dependence in the most aggressive regime.

\section{Conclusion}

This work presented a sample-efficient online model-based framework for precise, high-speed excavator control from scratch. Precision-gated contouring conditions rewarded progress on path accuracy, reshaping the conventional accuracy--progress trade-off to prioritize precision. Cross-evaluation showed that the learned dynamics support switching among the tested objectives at deployment without retraining. The controller learned directly on an 11.5-ton excavator, with all objectives reaching sub-centimeter mean contour error after 40 minutes.
Compared with the closest available learned excavator controllers, the method required less real-machine interaction, achieved comparable speed-normalized error, and reached higher operating speeds. Together, these results support online model-based learning for full-scale hydraulic machinery and other nonlinear, difficult-to-model systems.

A key direction for future work is to extend the framework beyond free-space motion to contact-rich tasks, incorporating force feedback and accounting for varying payloads and tool--soil interactions. An open question is how model-uncertainty estimates, such as ensemble disagreement, can guide exploration, support out-of-distribution detection, and inform risk-sensitive planning when learning these tasks from scratch. A complementary direction is to integrate environmental perception, collision-aware planning, and explicit safety constraints into the framework for deployment in unstructured environments.

\bibliographystyle{IEEEtran}
\bibliography{bibliography/references, bibliography/zotero_export_fn}

\newpage

\vfill

\end{document}